\documentclass[10pt,twocolumn,letterpaper]{article}

\usepackage{iccv}
\usepackage{times}
\usepackage{epsfig}
\usepackage{graphicx}
\usepackage{amsmath}
\usepackage{amssymb}
\usepackage{booktabs}
\usepackage{multirow}

\usepackage{graphicx,mathtools,kantlipsum}
\usepackage{longtable}
\usepackage{tablefootnote}

\usepackage{caption}
\usepackage{subcaption}

\usepackage[pagebackref=true,breaklinks=true,letterpaper=true,colorlinks,bookmarks=false]{hyperref}

\iccvfinalcopy 

\usepackage[capitalize]{cleveref}
\crefname{section}{Sec.}{Secs.}
\Crefname{section}{Section}{Sections}
\Crefname{table}{Table}{Tables}
\crefname{table}{Tab.}{Tabs.}
\usepackage{pifont}
\renewcommand{\paragraph}[1]{\noindent\textbf{#1}~~}

\usepackage[misc]{ifsym}
\newcommand\blfootnote[1]{\begingroup\renewcommand\thefootnote{}\footnote{#1}\addtocounter{footnote}{-1}\endgroup}

\begin{document}

\title{DORT: Modeling Dynamic Objects in Recurrent for Multi-Camera\\ 3D Object Detection and Tracking}

\author{Qing Lian\textsuperscript{1,2}
~~~~~Tai Wang\textsuperscript{1,3}
~~~~~Dahua Lin\textsuperscript{1,3}
~~~~~Jiangmiao Pang\textsuperscript{1\textrm{\Letter}} \\
\textsuperscript{1}Shanghai AI Laboratory
~~~\textsuperscript{2}The Hong Kong University of Science and Technology \\
\textsuperscript{3}The Chinese University of Hong Kong
 \\
{\tt\small qlianab@connect.ust.hk, \{wt019,dhlin\}@ie.cuhk.edu.hk, pangjiangmiao@gmail.com}}
\maketitle

\begin{abstract}

Recent multi-camera 3D object detectors usually leverage temporal information to construct multi-view stereo that alleviates the ill-posed depth estimation.
However, they typically assume all the objects are static and directly aggregate features across frames. 
This work begins with a theoretical and empirical analysis to reveal that ignoring the motion of moving objects can result in serious localization bias.
Therefore, we propose to model Dynamic Objects in RecurrenT (DORT) to tackle this problem. 
In contrast to previous global Bird-Eye-View (BEV) methods, DORT extracts object-wise local volumes for motion estimation that also alleviates the heavy computational burden. 
By iteratively refining the estimated object motion and location, the preceding features can be precisely aggregated to the current frame to mitigate the aforementioned adverse effects. 
The simple framework has two significant appealing properties.
It is flexible and practical that can be plugged into most camera-based 3D object detectors.  
As there are predictions of object motion in the loop, it can easily track objects across frames according to their nearest center distances. 
Without bells and whistles, DORT outperforms all the previous methods on the nuScenes detection and tracking benchmarks with 62.5\% NDS and 57.6\% AMOTA, respectively. The source code is available at \url{https://github.com/SmartBot-PJLab/DORT}. 

\blfootnote{\textrm{\Letter} Corresponding author.}

\end{abstract}

\section{Introduction}

\begin{figure}
    \centering
    \includegraphics[width=0.98\columnwidth]{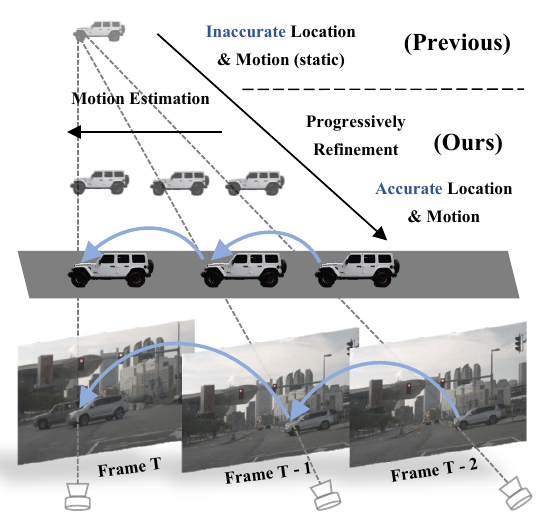}
    \caption{Visualization of object localization from temporal correspondence. 
    Previous work ignores the motion of moving objects, which leads to imprecise localization. Our work progressively refines the object's location and motion so that the preceding features can be precisely aggregated.}
    
    \label{fig:fig1}
    \vspace{-5mm}
\end{figure}

Multi-camera 3D object detection is critical to robotic systems such as autonomous vehicles. 
As object depth estimation from a single image is naturally ill-posed, 
recent works use large-scale depth pre-trained models~\cite{dd3d} and leverage geometric relationships~\cite{RTM3D,MonoFlex,wang2021PGD,wang2022dfm} to alleviate the problem.
Because stereo correspondence exists in consecutive frames, some works resort to temporal information for accurate depth predictions. 
For example, BEVDet4D~\cite{huang2021bevdet4d} and BEVFormer~\cite{li2022bevformer} warp preceding features to the current frame to enrich the single-frame BEV representations. 
DfM~\cite{wang2022dfm} constructs temporal cost volumes that explicitly establish the stereo correspondence.
However, these cross-frame feature aggregations do not consider the motion of moving objects and assume all the objects are static, which results in serious 3D localization bias.  

In this paper, we first provide a theoretical and empirical analysis to reveal the negative effects of inaccurate object motion to object depth (Fig~\ref{fig:fig1}).
In particular, if the object is moving toward or backward to the ego-car, the incorrect correspondence would derive a farther or closer depth.
In the driving scenarios, it is critical that a misleading farther depth is estimated, which might reduce the reaction time of the decision system, leading to catastrophic collision accidents. 
This motivates us to devise an explicit mechanism to involve object motion estimation in the temporal-based 3D object detection pipeline.

Modeling dynamic objects in this context has several challenges: (1) We need a flexible object-wise representation for potential object-wise operations based on motion modeling.
(2) Jointly estimating object location and motion is an inherent chicken and egg problem~\cite{Hartley_mvg}:
The temporal correspondence can derive accurate object location only when accurate object motion is given and vice versa.
(3) Simultaneously predicting object location and motion from only two frames is also an ill-posed problem theoretically, and thus involving right-body assumption and more frames in the framework to pose reasonable constraints is desired.

To address these problems, we model \textbf{D}ynamic \textbf{O}bjects in \textbf{R}ecurren\textbf{T} (DORT) that simultaneously estimates object motion and location, and then progressively refines them for accurate 3D object detection. 
It benefits from a local 3D volume representation that not only extracts object-wise 3D features but also alleviates the heavy computational costs of global BEV in previous methods~\cite{wang2022dfm, li2022bevformer, huang2021bevdet}.
Based on the object-wise 3D volume, temporal cost volumes are constructed by warping the volumes from the preceding frame to the current frame according to the object motion. 
Then the obtained cost volumes act as the features for updating the candidate location and motion. We model this estimation and update pipeline as a recurrent process to alleviate the aforementioned chicken and egg problem.
In addition, our framework can take into more than two frames and pose constraints to the object motion across different frames. It inherently provides a feasible solution to avoid the ill-posed dilemma of estimating object location and motion from only a single pair of correspondence observations.
As there is object motion prediction in the loop, the framework is naturally capable of joint object detection and tracking by
utilizing object motion to align the detection results into the same timestamp.
It also can be plugged into most camera-based 3D object detectors for flexible and practical use.

We validate the effectiveness of our framework on the nuScenes detection and tracking benchmarks. Benefiting from the dynamic objects modeling, DORT outperforms all
the previous methods with a large margin, leading to 62.5\% nuScenes detection metric (NDS) and 57.6\% and average multi-object tracking accuracy (AMOTA), respectively. 

\section{Related work}

\paragraph{Monocular 3D Object Detection}
Monocular-based 3D object detection was first approached from single-frame based methods and evolved into multi-frame based to alleviate the ill-posed depth estimation problem.

\noindent{\emph{(a) Methods with A Single Frame}}\quad
The single-frame based methods~\cite{zhou2019objects, MonoDLE, MonoFlex, MonoEF, brazil2019m3d} first extend 2D object detectors and insert several parallel regression heads (\textit{i.e.} 3D size, depth, and orientation) to predict 3D bounding boxes.
To alleviate the ill-posed depth recovery problem, several methods improve the model from the perspectives of loss module~\cite{MonoDLE}, network architecture~\cite{Brazil2020kinematic, wang2021fcos3d}, regression objective~\cite{MonoFlex, wang2021PGD}, etc.
Besides directly regressing object depth, later approaches~\cite{RTM3D, autoshape_liu, lian2022monojsg} further design 2D-3D geometry constraints to better extract visual cues for object depth estimation.
To align the detection features with output space, another line of methods~\cite{roddick2018orthographic} design several transformation modules to lift 2D inputs into 3D space. 
Pseudo-lidar based methods~\cite{pseudo_lidar, you2019pseudo, wang2019pseudo, wang2019pseudo} first predict the per-pixel depth and convert the raw pixel into point cloud for 3D detection.
BEV-based methods~\cite{roddick2018orthographic, philion2020lift, huang2021bevdet, li2022bevdepth} propose orthographic feature transformation (OFT) to transform the image features into a 3D voxel and then adopt a LiDAR-based head to localize objects.
Later work improves the OFT from the perspectives of explicit depth distribution modeling~\cite{philion2020lift, huang2021bevdet, li2022bevdepth}, incorporating deformable attention module~\cite{li2022bevformer} or designing 3D-based position encoding for attention~\cite{detr3d, liu2022petr}.

\noindent{\emph{(b) Methods with Multiple Frames}}\quad
Although many techniques are designed in single-frame based methods, they still suffer from ill-posed depth recovery problems, leading to unsatisfactory performance for deployment.
To augment the single-view observation, recent work~\cite{li2020jst, Brazil2020kinematic, wang2022dfm, huang2021bevdet4d, li2022bevformer, li2022bevstereo} leverages previous frames as additional observations for geometry modeling and features augmentation.
Kinematic3D~\cite{Brazil2020kinematic} leverages 3D Kalman Filter to associate objects across frames and refines the estimated 3D box. Later studies~\cite{wang2022dfm, huang2021bevdet4d, li2022bevformer} construct cross-frame cost-volumes as another visual cue for 3D detection. 
The cost volumes is based on the multi-view stereo, which assumes objects are static across frames. However, this assumption does not align with the driving scenario, where the objects can move. 
More critically, our analysis demonstrates that the inaccurate object motion would introduce misleading visual cues for object localization. 

\paragraph{Monocular 3D Object Tracking}
3D object tracking associates the objects across frames and generates a set of trajectories for motion prediction and planning. Traditional methods adopt a tracking-by-detection paradigm that first detects objects in each frame and then associates them by the appearance features~\cite{li2020jst} or objects' displacement with Kalman filter~\cite{Bewley2016_sort,yin2021center, Chen2022PolarDETR,Shi2022SRCN3D,Yang2022QualityTrack,fischer2022ccdt}. Besides the above paradigm, several methods~\cite{Hu2021QD3DT, Li_2022_time3d} design a two-stage paradigm that first associates objects based on appearance features and then utilizes the temporal motion to improve the object detection performance. 
In this work, we utilize temporal cost volumes to bridge the spatial location and temporal motion and derive a recurrent paradigm that iteratively updates them to obtain tightly coupled results for joint 3D detection and tracking.

\paragraph{Multi-View 3D Perception}
Leveraging multi-view images to recover 3D information is a fundamental topic, such as structure from motion~\cite{mvdepthnet}, multi-view stereo~\cite{yao2018mvsnet},  simultaneous localization and mapping~\cite{slam_review}, etc. 
One line of methods develop neural-network-based cost volumes~\cite{yao2018mvsnet, yao2019recurrent,sun2017pwc,zachary2020raft,zachary2021raft3d, sun2020disprcnn} to construct cross-frame visual cues for 3D perception. 
Another line of methods~\cite{Wang2019NOCS, tang2018banet,li2019multisensor} constructs geometry constraints and leverage optimization techniques to obtain a tight-coupled 3D structure. 
However, most of the work assumes the scene and objects are static, making them fail to handle the moving objects in driving scenarios.



\section{Object Motion in Temporal Modeling}

In this section, we first provide a theoretical and empirical analysis to demystify the adverse effects of neglecting object motion in temporal modeling~\cite{wang2022dfm, huang2021bevdet4d}, and then briefly discuss the challenges of modeling 3D motion in the monocular setting.

\subsection{Localization Bias from the Static Assumption}
In previous temporal-based methods~\cite{wang2022dfm, huang2021bevdet4d, li2022bevformer}, the object motion is ignored by assuming objects are static across frames and the temporal features are directly aggregated after converting the past frames to the current frame. We first show that the static assumption would derive a biased depth in temporal modeling.
Without loss of generality, we consider the two-view case here, and the derived analysis can be naturally extended to more than two views.
We denote the camera intrinsic as $K$ and the ego-motion from frame $t_0$ to frame $t_1$ as $T_{t_0\rightarrow t_1}^{ego}$:
\begin{align}
    K = \left[\begin{array}{cccc}
        f & 0 & c_u  \\
        0 & f & c_v \\
        0 & 0 & 1  \\
    \end{array}\right], 
    T_{t_0\rightarrow t_1}^{ego} = 
    \left[
    \begin{array}{cccc}
    1 & 0 & 0 & x^{ego} \\
    0 & 1 & 0 & 0 \\
    0 & 0 & 1 & z^{ego} \\
    \end{array}
    \right].
\end{align}
Here, $f$ is the camera's focal length, and $(c_u, c_v)$ is the camera center coordinates in the image. For simplicity, we assume the ego-motion only contains the translation $(x^{ego}, 0, z^{ego})$ on the horizontal plane. The analysis also can be easily extended to a more complicated case that the motion contains rotation.

Given the multiple-view images, temporal-based methods can utilize photometric or featuremetric similarity to find the correspondence of pixel $p_{t_0} = (u_{t_0}, v_{t_0})$ in the past frame $t_0$ and the pixel $p_{t_1} = (u_{t_1}, v_{t_1})$ in the current frame $t_1$. The depth $z_{t_1}$ can be recovered by solving the correspondence:
\begin{align}
    \label{eq:nomotion}
    & T_{t_0\rightarrow t_1}^{ego}\cdot \pi(p_{t_0}, K) = \pi(p_{t_1}, K), \\
    & z_{t_1} = \frac{z^{ego}(u_{t_0} - c_u) - fx^{ego}}{u_{t_0} - u_{t_1}}, 
\end{align}
where $\pi$ denotes the projection from 2D image coordinate to 3D camera coordinate.
The above derivation assumes that the object is static across two frames. However, in the driving scenario, objects can move with a corresponding object motion $T^{obj}_{t_0 \rightarrow t_1}$.
Suppose the object motion only contains the translation on the horizontal plane in a short time interval, it can be represented as:
\begin{align}
    T^{obj}_{i \rightarrow j} = \left[
        \begin{array}{cccc}
            1 & 0 & 0 & x^{obj} \\
            0 & 1 & 0 & 0 \\
            0 & 0 & 1 & z^{obj} \\
        \end{array}
    \right].
\end{align}
Then the corresponding equation to recover depth can be revised as follows:
\begin{align}
    \label{eq:motion}
    & T^{obj}_{t_0 \rightarrow t_1}T_{t_0\rightarrow t_1}^{ego} \cdot \pi(p_{t_0}, K) = \pi(p_{t_1}, K),\\
    & \hat{z}_{t_1} = \frac{(z^{ego} + z^{obj})(u_{t_0} - c_u) - f(x^{ego} + x^{obj})}{u_{t_0} - u_{t_1}}.
\end{align} 
Based on Eq~\eqref{eq:nomotion} and \eqref{eq:motion}, we can obtain the depth gap:
\begin{align}
    \Delta z = \frac{z^{obj} (u_{t_0} - c_u) - fx^{ego}}{u_{t_0} - u_{t_1}}.
    \label{eq:bias_depth}
\end{align}




\begin{figure}[t]
    \centering
    \includegraphics[width=\columnwidth]{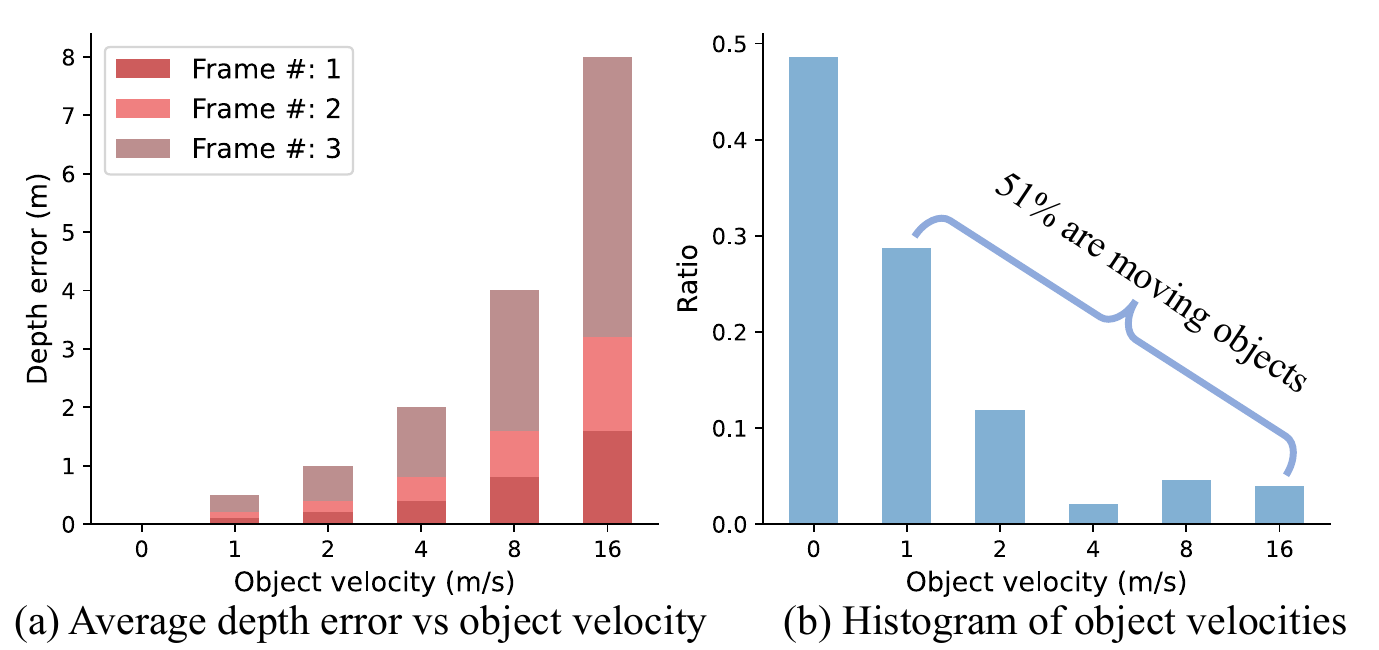}
    \caption{Empirical analysis of the depth bias on the nuScenes dataset if objects are assumed static.}
    \vspace{-4mm}
    \label{fig:analysis}
\end{figure}

\begin{figure*}[htb]
    \centering
    \includegraphics[width=\textwidth]{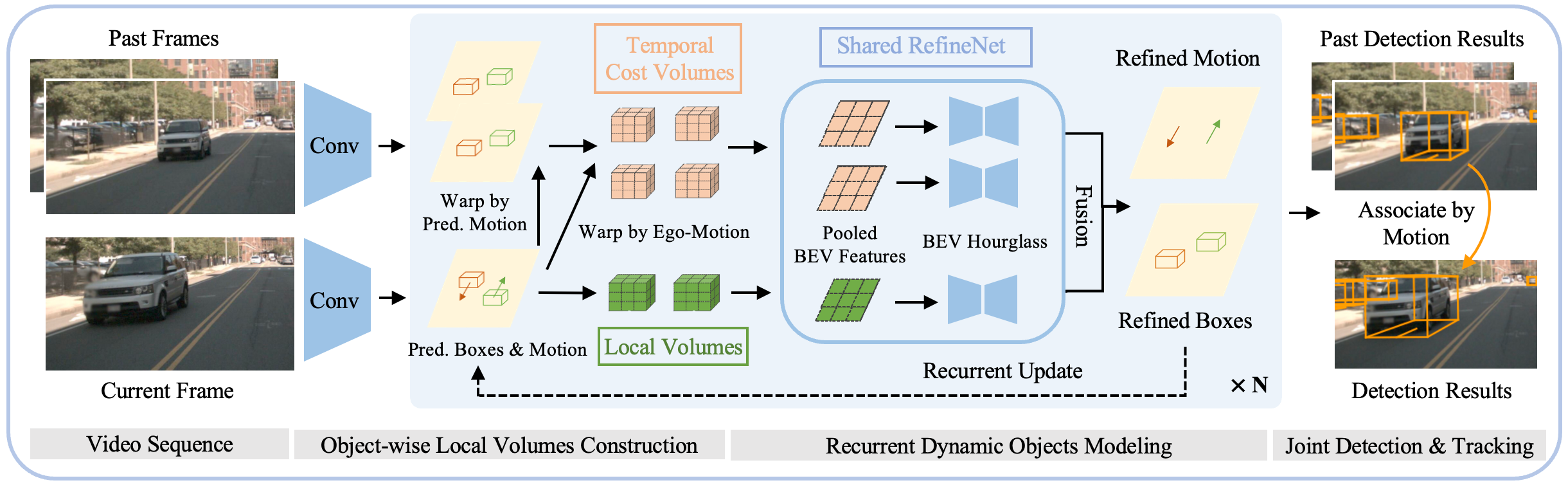}
    \caption{Pipeline overview.  Given a video sequence, we first extract the features from 2D images and generate the candidate boxes and their motion by a single-frame detector. Then the boxes and motion are progressively refined from the concurrently updated corresponding 3D volume features. A fusion process in the recurrent module finally combines the estimation from each pair of frames. Based on the tightly coupled modeling of object location and motion, the framework can achieve joint 3D detection and tracking during inference. }
    \label{fig:pipeline}
    \vspace{-4mm}
\end{figure*}

From Eq~\eqref{eq:bias_depth}, we can observe that the depth bias is linearly correlated with the object motion. In Fig~\ref{fig:analysis}, we also display the empirical statistics of object motion and the corresponding depth bias from the nuScenes dataset. We can observe that the empirical depth error is also linearly correlated to the object velocity and increases as the time interval enlarges. Besides, the right part in Fig.~\ref{fig:analysis} also shows that almost 51\% of objects are moving across frames, demonstrating the necessity of modeling object motion in the temporal-based framework.

\subsection{Ill-Posed Problem in Motion Modeling}
Except for demonstrating the necessity of modeling object motion in temporal-based frameworks, we also want to mention that simultaneously estimating object location and motion is a non-trivial problem, especially in the two-frame case. As shown in Fig~\ref{fig:fig1}, the correspondence of two points from the temporal frames can come from infinite combinations of object location and motion (We provide more examples and the visualization in the supplementary).
This illustrates that joint estimation of location and motion from only one correspondence is an ill-posed problem. 
To alleviate this issue, we first simplify the object motion as a right-body movement so that multiple correspondences from the points in the object can be used to solve a shared motion.
Furthermore, we also leverage more than two frames to constrain the flexibility of object motion. More details can be referred to Sec~\ref{sec:fusion}.

\section{Methodology}
This section describes the details of DORT. 
DORT is a general joint detection and motion prediction module that can estimate coupled object location and motion results across frames. Based on the tightly coupled location and motion results, DORT is also capable of simultaneously 3D object detection and tracking. Basically, it can be based on most temporal 3D detectors~\cite{huang2021bevdet4d, park2023time, li2022bevformer}. In this work, we select the popular temporal detector BEVDepth~\cite{li2022bevdepth} as the base detector and extend it to handle both static and moving objects in temporal modeling.
We first present an overview of temporal-based frameworks in Sec~\ref{sec:dfm} and then introduce our modifications: 
the local volume for object-wise representation in Sec~\ref{sec:local_volume},
the key recurrent dynamic objects modeling in Sec.~\ref{sec:recurrent},
and the object association for monocular 4D object detection in Sec~\ref{sec:track}.

\subsection{Overview of Temporal-Based Frameworks}
\label{sec:dfm}
Previous temporal-based camera-only 3D detection methods contain three stages: 
(1) The 2D features extraction stage extracts the features from the input images.
(2) The view transformation and stereo matching process that first lift the 2D features to a global 3D volume and then warp the features in each frame to an aligned canonical space for stereo matching. Depending on the model design, the view transformation and stereo matching order may reverse.
(3) The detection stage fuses the 3D features from different observations (monocular and stereo) and estimates 3D bounding boxes based on a BEV-based detection head.

In this work, we follow previous methods~\cite{wang2022dfm, huang2021bevdet, li2022bevdepth} and adopt the widely-used 2D backbone (\textit{e.g.} ResNet~\cite{he2016deep}) to do the features extraction.
For the view transformation stage, we design an \textbf{object-wise local volume} that leverages the candidate 3D boxes to obtain the potential foreground regions and only models them with local object-wise 3D volumes.
For the stereo matching and detection stages, we propose a \textbf{recurrent dynamic objects modeling} module. Specifically, we adopt a recurrent paradigm that progressively refines the detection and motion results by concurrently updating the corresponding 3D volume features thereon.

\begin{figure}
    \centering
    \includegraphics[width=0.44\textwidth]{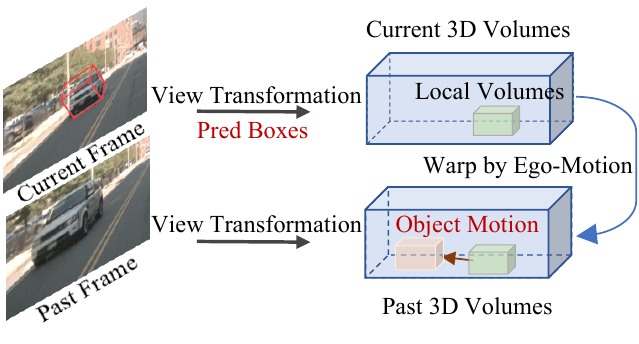}
    \vspace{-2mm}
    \caption{The process of extracting local volumes in the current and past frames according to predicted bounding boxes and object motion.}
    \vspace{-5mm}
    \label{fig:local}
\end{figure}

\subsection{Object-wise Local Volume}
\label{sec:local_volume}
In previous work, the 2D-3D transformation considers each candidate 3D grid point and constructs a global volume for 3D detection. However, there are several limitations:
(1) the global 3D volume contains lots of background regions, which is not vital for 3D detection but largely increases the computation burden.
(2) Modeling a global volume needs to pre-define a detection range during training, making the detectors fail to detect objects with arbitrary depths.
(3) It is inconvenient to maintain and manipulate a global 3D volume with various object-wise operations.

Hence, we replace the global 3D volume with an object-wise local volume. Specifically, we leverage the candidate bounding boxes to determine the 3D region of interest (RoI) and set the local volume center as the bounding box center.
To keep the object ratio and achieve cross-view warping, we assign each 3D RoI volume $V \in \mathcal{R}^{W\times H\times L\times C}$ with the same 3D dimension $(W, H, L)$ and channel size $C$.  
Different from 2D detection, the objects' dimension in 3D space has less variance and empirically relies less on the RoI-Align~\cite{he2017mask} operation.
We display the construction of object-wise local volumes in Fig~\ref{fig:local}.
For the 2D to 3D transformation, we first follow LiftSplat~\cite{philion2020lift} and lift the images to a 2.5D frustum by weighting with depth probability. Then we utilize the grid sample operation to warp the features from the 2.5D frustum to each 3D local volume. 
Benefiting from the accurate 2D detection performance of perspective-view detection, the local volume features sampled from the 2.5D frustum would have a large overlap with the foreground objects. Hence, if the proposal 3D location is inaccurate, the later refinement module still has the ability to use the features to do the refinement.

\subsection{Recurrent Dynamic Objects Modeling}
\label{sec:recurrent}

The pipeline of the recurrent framework is illustrated in Fig.~\ref{fig:pipeline}. 
Given the candidate 3D bounding boxes and motion as input, each iteration first constructs the temporal cost volumes thereon, and aggregates these cues to refine the proposal boxes and motion. 
In particular, we adopt a perspective-view based 3D detector (\textit{i.e.} PGD) to generate the initialized candidate 3D boxes and motion and only predict their residuals for refinement in the subsequent recurrent updates. Next, we elaborate on these steps in detail.

\begin{table*}
\centering
\caption{Experimental results of monocular 3D object detection and tracking on the nuScenes test set. The input resolution is $1600\times 900$ with using ConvNeXt-Base~\cite{liu2022convnet} as the backbone.}
\label{tab:main_test}
\begin{subfigure}{0.56\textwidth}
\makeatletter\def\@captype{table}
    
\centering
\caption{3D detection results on the nuScenes test set.}

\vspace{-2mm}
\label{tab:main_test_set_det}

\setlength{\tabcolsep}{1pt}
\resizebox{0.95\linewidth}{!}{\begin{tabular}{l|ccccccccc} 
\hline
Method    & mAP$\uparrow$   & mATE$\downarrow$ & mASE$\downarrow$   &mAOE$\downarrow$   &mAVE$\downarrow$   &mAAE$\downarrow$   &NDS$\uparrow$\\\hline
Ego3RT~\cite{lu2022ego3rt} & 42.5  & 0.55 & 0.26 & 0.43 & 1.01 & 0.14 & 47.3 \\
UVTR~\cite{li2022uvtr} & 47.2 & 0.57 & 0.25 & 0.39 & 0.51 & 0.12 & 55.1  \\ 
BEVFormer~\cite{li2022bevformer} & 48.1 & 0.58 & 0.25 & 0.37 & 0.37 & 0.12 & 56.9  \\
PETRv2~\cite{liu2022petrv2} & 51.2  & 0.55 & 0.25 & 0.36 & 0.40 & 0.13 & 58.6 \\
BEVDepth~\cite{li2022bevdepth} & 52.0  & 0.45 & 0.24 & 0.35 & 0.35 & 0.13 & 60.9  \\
BEVStereo~\cite{li2022bevstereo} & 52.5  & 0.43 & 0.24 & 0.36 & 0.35 & 0.14 & 61.0\\ 
SOLOFusion~\cite{park2023time} & 54.0  & 0.45 & 0.26 & 0.37 & 0.27 & 0.14 & 61.9\\ \hline
DORT (Ours) & \textbf{54.5} & 0.44 & 0.26 & 0.37 & \textbf{0.25} & 0.14 & \textbf{62.5}\\  \hline
\end{tabular}}


\end{subfigure}
 \hspace{0.1cm}
 \begin{subfigure}{0.37\textwidth}
\makeatletter\def\@captype{table}
    

\caption{3D  tracking results on the nuScenes test set.}
\label{tab:main_test_set_track}
\centering
\setlength{\tabcolsep}{2pt}
\resizebox{0.95\linewidth}{!}{\begin{tabular}{l|cccc} 
\hline
Method     & AMOTA$\uparrow$  &AMOTP$\downarrow$  & MOTAR $\uparrow$  \\\hline
QD-3DT~\cite{Hu2021QD3DT} & 21.7 & 1.550 & 56.3 \\
Time3D~\cite{Li_2022_time3d} & 21.4 & 1.360 & - \\
PolarDETR~\cite{Chen2022PolarDETR}         & 27.3 & 1.185 & 60.7 \\ 
MUTR3D~\cite{zhang2022mutr3d}           & 27.0 & 1.494 & 64.3 \\ 
SRCN3D~\cite{Shi2022SRCN3D} & 39.8 & 1.317 & 70.2 \\
QTrack~\cite{Yang2022QualityTrack} & 48.0 & 1.100 & 74.7 \\
UVTR~\cite{li2022uvtr} & 51.9 & 1.125 & 76.4 \\ \hline
DORT(Ours) & \textbf{57.6} & \textbf{0.951} & \textbf{77.1} \\  \hline
\end{tabular}}
\end{subfigure}
\vspace{-3mm}

\end{table*}

\subsubsection{Cross-Frame Cost Volumes Construction}
\label{sec:cost_volume}
Given the initial predictions of 3D boxes and their motion, we first obtain object-wise volume features following Sec.~\ref{sec:local_volume}. 
Then we can construct the temporal cost volumes by warping features from past frames to the current frame coordinates based on \emph{ego-motion}. 
In contrast to previous work~\cite{wang2022dfm,huang2021bevdet4d} assuming objects are static, we further involve the \emph{object motion} into the warping procedure.
Specifically, for each point $p \in \mathcal{R}^{3}$ in the object-wise local volume $V$, we query the corresponding features in previous frame $t - \Delta t$ with the consideration of ego-motion $T^{ego}$ and the object motion $T^{obj}$ and construct the cost as $\left[V(p), V_{t - \Delta t}(T^{obj}T^{ego}p) \right]$.
Note that we simplify the point motion as the object motion with a rigid-body assumption, which can approximate most of the cases in driving scenarios, especially for vehicles~\cite{li2020jst, zachary2021raft3d}.

\subsubsection{3D Boxes and Motion Residual Estimation}
Given the object-wise temporal features built from input 3D boxes and motion, we leverage a refinement network to estimate the residual between the input 3D boxes and motion with the ground truth. 
The refinement network contains several 2D/3D residual-based convolutional layers to extract the 3D volumes and 2D BEV features. 
The detailed architecture is presented in Supplementary. 
Formally, the refinement is formulated as the regression of 3D attribute residuals $\mathcal{B}$, including the object's 3D center $x, y, z$, 3D size $w, h, l$, rotation $\theta$ and velocity $v_x, v_y$.
Since we use the object velocity in the current frame to represent the object motion and assume constant velocity across frame, the supervision for some frames may contain noise(\textit{e.g.} inaccurate labels, violation of rigid-body assumption, etc). Hence, we model the residual as a Laplacian distribution and design the loss function as:
\begin{align}
    \label{eq:depth}
    \mathcal{L}_{refine} = \sum_{b\in\mathcal{B}}(\frac{\sqrt{2}}{\sigma_b} \|\Delta\hat{b} - \Delta b\| + \log{\sigma_b}).
\end{align}
Here, $\Delta b$, $\Delta\hat{b}$, and $\sigma_b$ are all the network outputs, and represent the ground truth residual, the estimated residual, and the estimated standard deviation of residual for each 3D attribute, respectively.

\subsubsection{Multiple Estimation Fusion}
\label{sec:fusion}
Given $n$ frame as inputs, we can obtain $n$ 3D volumes ($1$ for the local volumes from the referenced single view and $n-1$ for the paired cross-view cost volumes) and obtain $n$ estimated residuals from the above residual estimation module. Then we weigh the importance of each residual by the estimated deviation and fuse them to obtain an ensemble result.
The fusion formulation can be represented as:
\begin{align}
    \hat{b}_{fused} = \sum_{i=1}^{n}\frac{e^{\sigma_{b_i}} b_i}{\sum_{i=1}^{n}e^{\sigma_{b_i}}},
\end{align}
where $i$ denotes the volume index. 

For simplicity, here we only estimate the velocity measurement for the referenced frame, \emph{i.e.}, the fluctuation of object velocity across different frames would not be considered explicitly. The mechanism for multi-frame fusion is expected to handle this problem adaptively. At the same time, this constraint also provides additional cues when simultaneously estimating object location and motion from more than two frames.

\subsubsection{Recurrent Location and Motion Update}
After each iteration, we can obtain the refined bounding boxes with their motion and thus can derive the updated bounding boxes in different frames. With these updated locations and RoIs, we can further update the volume features and proceed to the next-round refinement. Note that any complex or learnable motion modeling can be integrated into this procedure. Here, to be consistent with the multiple estimation fusion designs, we still keep the constant velocity prediction for simplicity to derive the bounding boxes of previous frames. 

In the training stage, we follow the recurrent methods~\cite{zachary2020raft, zachary2021raft3d} in other tasks and set the loss weight for each iteration as the same.
The overall loss is represented as:
\begin{align}
    \mathcal{L} = \mathcal{L}_{pv} + \sum_{i=1}^{k} \mathcal{L}^{i}_{refine},
\end{align}
where $ \mathcal{L}_{pv}$ is the loss in the perspective-view detector~\cite{wang2021PGD}, $\mathcal{L}^i_{refine}$ is the refinement loss in each iteration and $k=3$ is the number of the iterations.

In the inference stage, we first take the perspective-view detector (\textit{i.e.} PGD~\cite{wang2021PGD}) to generate the initial 3D bounding boxes and their motion and then progressively refine them with three iterations. During each iteration, we first construct the volume features as discussed in Sec~\ref{sec:cost_volume} and feed them into the recurrent refinement module to estimate the 3D boxes and motion residuals for each paired frame input. Then we utilize the multiple estimation fusion module in Sec~\ref{sec:fusion} to fuse the estimated results and obtain the refined 3D boxes and motion as the next stage input. 
\subsection{Monocular 4D Object Detection}
\label{sec:track}
So far, we have introduced our recurrent framework for 3D detection from monocular videos.
Based on progressive refinement, our model can estimate tightly coupled object location and motion results and thus can easily associate the object detection results across frames, leading to joint 3D detection and tracking.
Specifically, we follow~\cite{Bewley2016_sort, zhou2020tracking, yin2021center, pang2021simpletrack} and associate the detection results by warping current detection to the past frames with object motion. 
Based on the ego-motion, we first convert the predicted object location to the past frame coordinate and then warp with the estimated object velocity. Then we follow the popular distance-based tracker in 2D task~\cite{Bewley2016_sort, yin2021center} and associate the objects by the closest distance matching. We also utilize the Kalman filter to maintain the trackers' location and velocity.
We provide more details of the tracking pipeline in the supplemental materials.

\begin{table*}[t]
    \centering

    \caption{Experimental results on the nuScenes validation set. The input resolution is $704\times 256$ using ResNet-50 as the backbone.
    * denotes the re-implementation based on the provided code.}
    \vspace{-2mm}

    \resizebox{0.75\linewidth}{!}{
    \begin{tabular}{l|c|ccccccc} \hline
    \textbf{Methods}  & \# frame & \textbf{mAP}$\uparrow$  & \textbf{mATE}$\downarrow$ & \textbf{mASE}$\downarrow$   &\textbf{mAOE}$\downarrow$   &\textbf{mAVE}$\downarrow$   &\textbf{mAAE}$\downarrow$ & \textbf{NDS}$\uparrow$  \\ \hline
    PGD*~\cite{wang2021PGD} & \multirow{4}*{1}   & 28.8  & 0.75 & 0.27 & 0.52 & 1.13 & 0.18 & 37.0   \\
    BEVDet~\cite{huang2021bevdet} & & 29.8& 0.73 & 0.28 & 0.59 & 0.86 & 0.24  & 37.9   \\
    PETR~\cite{liu2022petr}& & 31.3  & 0.77 & 0.28 & 0.56 & 0.92 & 0.23 & 38.1    \\
    DETR3D~\cite{detr3d} & & 34.9  & 0.72 & 0.27 & 0.38 & 0.84 & 0.20 & 43.4\\ \hline
    BEVDet4D~\cite{huang2021bevdet4d}  & \multirow{3}*{2} & 32.2 & 0.70 & 0.28 & 0.50 & 0.35 & 0.21 & 45.7 \\
    BEVDepth~\cite{li2022bevdepth} & & 35.1  & 0.64 & 0.27 & 0.48 & 0.43 & 0.20 & 47.5\\
    DORT (Ours) & & \textbf{37.9}& 0.62 & 0.27 & 0.35 & \textbf{0.31} & 0.20 & \textbf{51.4}   \\ \hline
    BEVDepth*~\cite{huang2021bevdet4d} &  \multirow{2}*{8} & 39.8 & 0.57 & 0.27 & 0.49 &0.27  & 0.18  & 52.3 \\
    DORT (Ours)& & \textbf{41.8}  & 0.57 & 0.26 & 0.43 & \textbf{0.25} & 0.19 & \textbf{53.4} \\ \hline
    SOLOFusion~\cite{park2023time}& \multirow{2}*{16}  & 42.7 & 0.57 & 0.27  & 0.41  & 0.25 &  0.18  & 53.4 \\
    DORT (Ours) & &  \textbf{43.6}  & 0.56 & 0.26 & 0.41 & \textbf{0.24} & 0.18 & \textbf{54.0} \\ \hline
    \end{tabular}}
    \label{tab:nusc_val}
    \vspace{-3mm}
    \end{table*}


\begin{table}[htb]
    \centering
    \caption{3D object tracking results on the nuScenes validation set. We adopt ResNet-50 as the backbone and set the input resolution as $704\times 256$.}
    \vspace{-3mm}
\resizebox{0.8\linewidth}{!}{
    \begin{tabular}{c|ccc} \hline
         Method & AMOTA$\uparrow$ & AMOTP$\downarrow$ &   Recall$\uparrow$  \\ \hline
         QD-Track3D~\cite{Hu2021QD3DT} & 24.2 & 1.518 & 39.9 \\
         Time3D~\cite{Li_2022_time3d} &  21.4 & 1.360 & N/A \\ 
         TripletTrack~\cite{Marinello2022CVPR} & 28.5 & 1.485 & N/A \\ 
         MUTR3D~\cite{zhang2022mutr3d} & 29.4 & 1.498 & 42.7 \\ 
         QTrack~\cite{Yang2022QualityTrack} & 34.7 & 1.347 & 46.2 \\ \hline
         DORT & \textbf{42.4} & \textbf{1.264} & 49.2 \\  \hline
    \end{tabular}}
    \label{tab:nusc_track_val}
    \vspace{-5mm}
\end{table}

\section{Experiments}
We validate the effectiveness of our method on the large-scale nuScenes detection and tracking benchmark~\cite{nuscene}.

\subsection{Experimental Setup}
In this section, we describe the used dataset, the evaluation metrics, and the implementation details.

\noindent{\textbf{Dataset}}\quad
NuScenes~\cite{nuscene} is a large-scale autonomous driving dataset, which contains 1,000 video sequences. 
The official protocol splits the video sequences into 700 for training, 150 for validation, and 150 for testing. 
Each sequence is annotated with 3D bounding boxes, object velocity, and the tracking id to connect the objects across the frame. 
The camera calibration information, timestamps, and ego-motion in both training and test sets are provided.

\noindent{\textbf{Detection Metrics}}\quad
We adopt the official evaluation protocol provided by nuScenes benchmark~\cite{nuscene}. 
The official protocol evaluates 3D detection performance by the metrics of average translation error (ATE), average scale error (ASE), average orientation error (AOE), average velocity error (AVE), and average attribute error (AAE).
Besides, it also measures the mean average precision (mAP) with considering different recall thresholds. Instead of using 3D Intersection over Union (IoU) as the criterion, nuScenes defines the match by 2D center distance $d$ on the ground plane with thresholds $\{0.5, 1, 2, 4\}m$.
The above metrics are finally combined into a nuScenes Detection Score (NDS).

\noindent{\textbf{Tracking Metrics}}
Regarding the tracking metrics, the nuScenes benchmark mainly measures the average multi-object tracking accuracy (AMOTA), average multi-object tracking precision (AMOTP), and tracking recall. In particular, AMOTA and AMOTP are the averages of multi-object tracking accuracy (MOTA) and multi-object tracking precision (MOTP) under different recall thresholds. 

\noindent{\textbf{Network Details}}
As discussed in Sec~\ref{sec:recurrent}, the recurrent module requires a proposal detector to generate candidate foreground regions as the $1^{st}$ stage input. We adopt the popular monocular 3D detector PGD~\cite{wang2021PGD} due to its high 2D object detection recall.
Following~\cite{huang2021bevdet, huang2021bevdet4d, li2022bevdepth}, we adopt the ResNet-50~\cite{he2016deep} with FPN as the 2D feature extractor and mainly conduct experiments on this setting. The 2D feature extractors in PGD and the recurrent module are shared to save computation time. The grid size of the 3D volume is set as $0.8m$ with the range of $[-5m, 5m]$ in the X and Z (depth) axis and $[-4m, 2m]$ in the Y (height) axis. 
During 2D to 3D features transformation, we follow~\cite{li2022bevdepth} and adopt the depth distribution guided 2D to 3D features lifting. 
Regarding the test set submission, we follow~\cite{li2022bevdepth, li2022bevstereo} and adopt the ConvNeXt-Base~\cite{liu2022convnet} as the image backbone. 
The image backbone is initialized with ImageNet pre-trained weights, and no other external data is used.
We provide more details about the network architecture of the recurrent module and the training configuration in the supplementary materials.


\noindent{\textbf{Training Configurations}}
The model is optimized by AdamW optimizer with weight decay $10^{-2}$.
We first follow~\cite{wang2021PGD} to train the proposal detector and refine the recurrent module with 24 epochs, where the initial learning rate is set as $2\times 10^{-4}$ and decreases to $2\times 10^{-5}$ and $2\times 10^{-6}$ at the $18^{th}$ and $22^{th}$ epochs. Following~\cite{huang2021bevdet, li2022bevdepth}, we use the class balance sampling strategy (CBGS) to alleviate the class imbalance problem. 
We adopt the commonly used 2D data augmentation that randomly flips the image, resizes the image with the range of $[0.36, 0.55]$, and crops the image to the resolution of $704 \times 256$.
Regarding the input video sequences, we follow~\cite{huang2021bevdet4d, li2022bevdepth} and sample the preceding keyframes to obtain the past video sequences.
Regarding the test set submission, we enlarge the input resolution to $1600 \times 640$ and reduce the volume size to $0.4m$.

\subsection{Main Results}
In Table~\ref{tab:main_test}, we first provide the comparison of our framework with existing state-of-the-art methods on the nuScenes test benchmarks. 
We draw the following observations: 
(i) Benefiting from dynamic objects modeling, our method displays a significant improvement in both object detection (mAP) and motion estimation (mAVE),  and 0.5\%  and relatively 7.4\% better than the previous best method. These localization and motion estimation improvements also contribute to state-of-the-art results in terms of the nuScenes detection metric (NDS).
(ii) With strong localization and motion estimation results, our tracking module can better associate the detected objects in different timestamps, resulting in superior performance over all the other trackers with different metrics.
Specifically, we improve the second best tracker~\cite{li2022uvtr}, another distance-based tracker with 10.9\% and 7.5\% relative improvements on the AMOTA and AMOTP metrics.  Compared with the joint detection and tracking methods QDTrack3D~\cite{Hu2021QD3DT} and Time3D~\cite{Li_2022_time3d}, the performance gain of our method demonstrates the effectiveness of our dynamic objects modeling framework in jointly modeling object motion and location. 
(iii) In Table~\ref{tab:nusc_val} and ~\ref{tab:nusc_track_val}, we also report our method on the nuScenes validation set with different settings. For the detection performance, we can draw the same observation as in the test set that our method can outperform previous temporal-based methods~\cite{huang2021bevdet4d, li2022bevdepth, park2023time} in terms of mAP and mAVE. 
Note that our method is also compatible with the components designed in the current BEV-based frameworks, such as the training techniques in BEVDepth~\cite{li2022bevdepth} and the depth estimation module in SOLOFusion~\cite{park2023time}. Furthermore, the local 3D volume is more friendly to practical applications, which can handle objects with arbitrary depth in the image.

\begin{figure}[!htb]
    \centering
    \subfloat[mAP in each iteration]{
    \includegraphics[width=0.45\columnwidth, angle=0,trim= 15 0 15 0, clip]{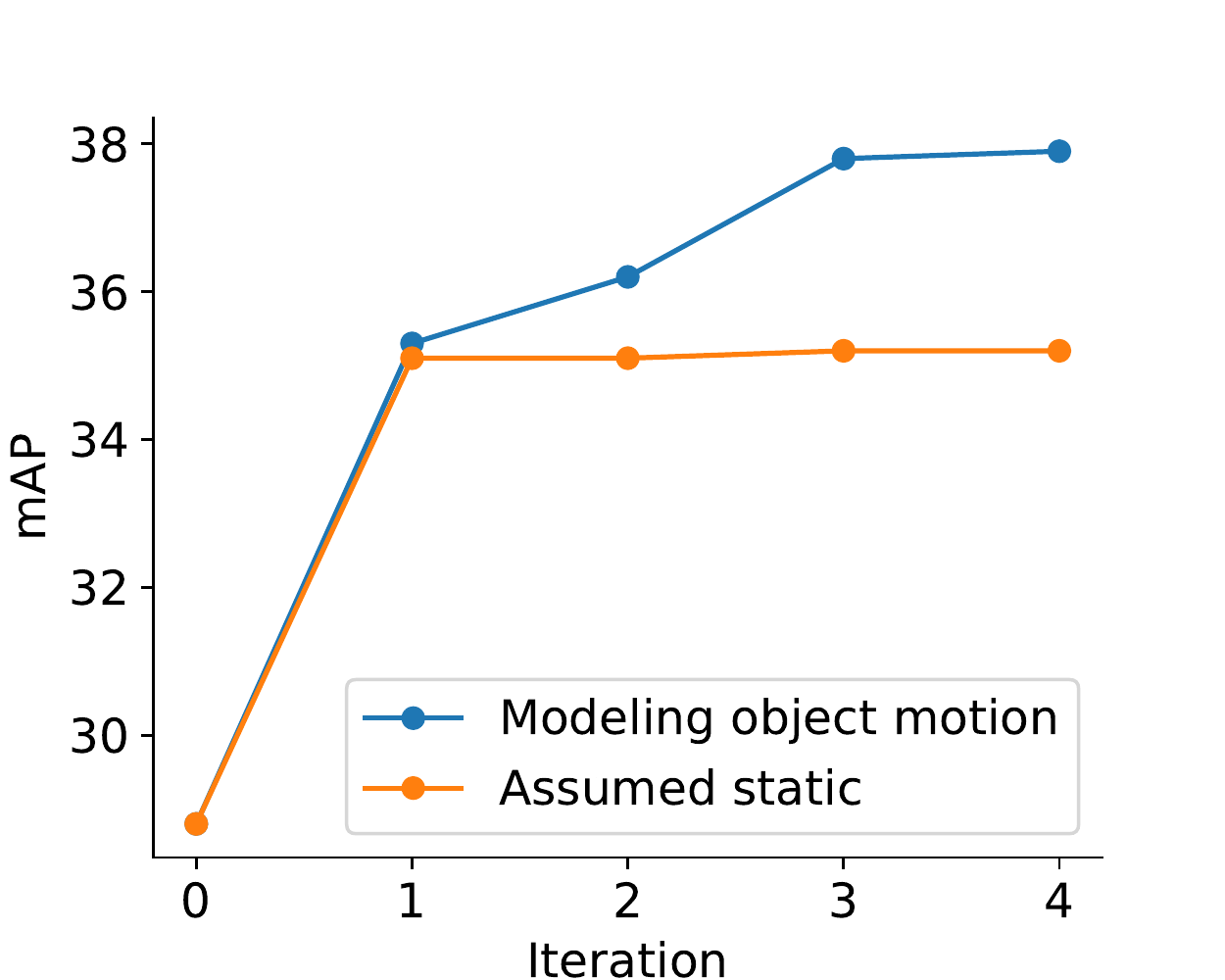}
    \label{fig:fig1a}
    }
    \quad
    \subfloat[mAVE in each iteration]{%
    \label{fig:fig1b}%
    \includegraphics[width=0.45\columnwidth, angle=0,trim= 15 0 15 0, clip]{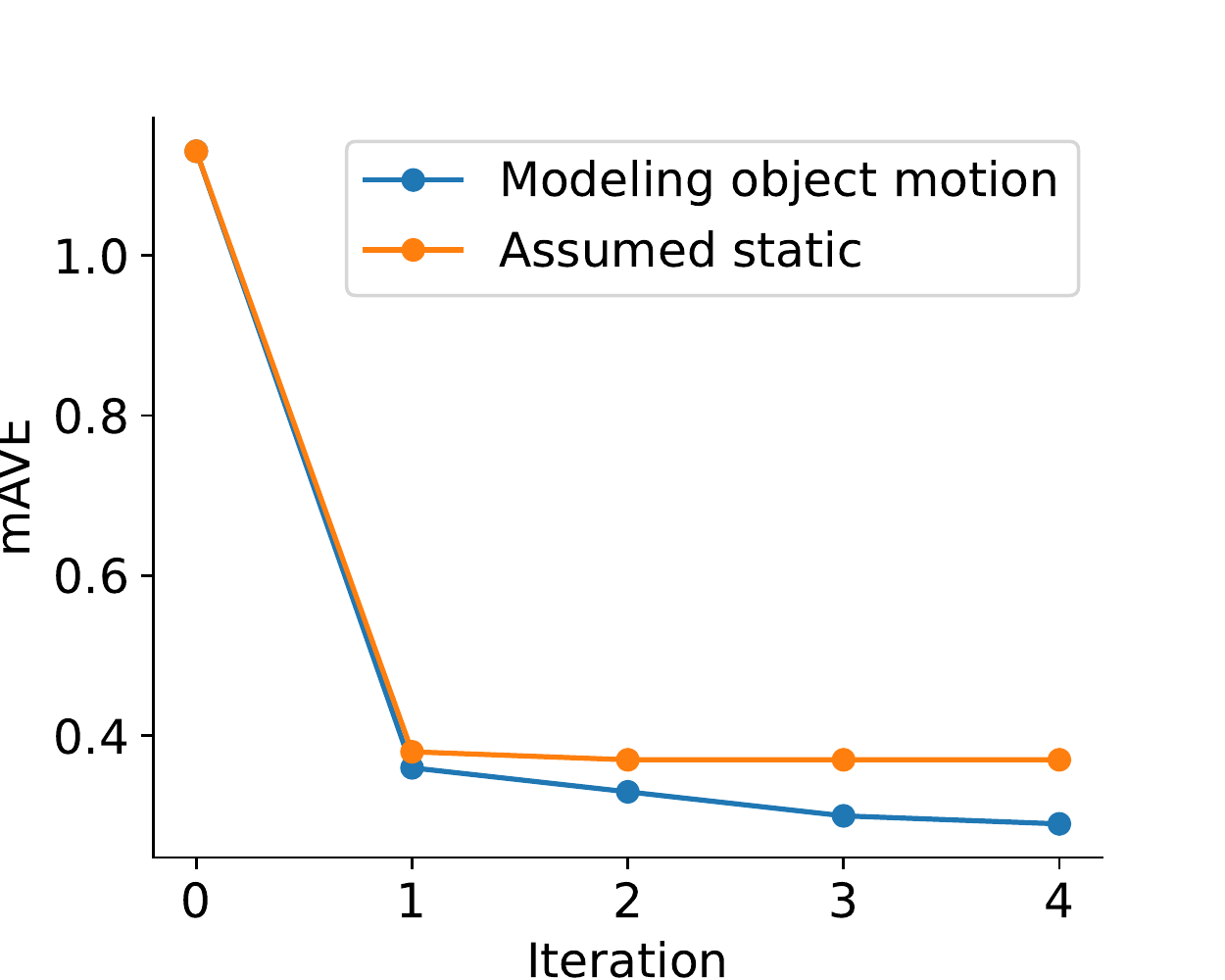}
    }%
    \caption{Detection results of each iteration in the recurrent pipeline. 1 past frame is used in temporal modeling.}
    \label{fig:iteration}
    \vspace{-5mm}
\end{figure}





\begin{table}[tb]
    \vspace{-3mm}
    \centering
    \caption{Ablation study of the local volumes on the nuScenes validation set.
     * denotes the single frame version without using EMA for fair comparisons. 
     The first term and the second term in FLOPS denote the operation in perspective view and bird-eye-view, respectively.}
\resizebox{0.85\linewidth}{!}{    \begin{tabular}{l|ccc} \hline
    Method & mAP & NDS & FLOPS \\ \hline
    PGD~\cite{wang2021PGD} & 28.8 & 37.0 & 238.2 + 0 \\ 
    BEVDepth~\cite{li2022bevdepth} & 33.3 & 40.6 & 120.4 + 74.4 \\ 
    PGD + Local Volume & 33.1 & 40.5 & 238.2 + 25.2 \\\hline
    \end{tabular}}
    \label{tab:local}
    \vspace{-5mm}
\end{table}

\subsection{Ablation Study}

\noindent{\textbf{Effectiveness of Object-wise Local Volume}}\quad
We first show the effectiveness of the object-wise local volume in the single-frame setting. Since there is no motion modeling and temporal volume construction, the experiments are conducted with one iteration in the recurrent step. As shown in Table~\ref{tab:local}, the object-wise local volume can improve the proposal detector PGD with a large margin and achieve comparable performance with BEVDepth, which exploits a global volume. In addition, the computation cost for the local volume is relatively small compared to the global 3D volume, enabling us to achieve multiple iterations and use more preceding frames to boost the performance. Note that the proposal detector also can be replaced by a more efficient one.

\noindent{\textbf{Ablation Study of Object Motion}}\quad
We further validate the influence of different dynamic object modeling strategies on the detection performance The first experiment compares the assumed static case with that of using the ground truth object motion. As shown in Table~\ref{tab:motion}, the model with ground truth object motion outperforms the assumed static with 4.3\% mAP, demonstrating the necessity of object motion for obtaining accurate temporal correspondence features. When we replace the ground truth object motion with an estimated one, it still can bring 2.9\% mAP improvements, illustrating the usefulness of our dynamic objects modeling module.

\begin{table}[tb]
    \centering
    \caption{Experimental results of different motion modeling strategies on the nuScenes validation set. The GT motion is obtained from the ground truth velocity. 1 past frame is used in temporal modeling.}
    \resizebox{0.7\linewidth}{!}{\begin{tabular}{c|ccc}\hline
    Setting & mAP & NDS & mAVE \\ \hline
   Assumed static & 35.0 & 47.1 & 0.37 \\
   GT motion & 39.3 & - & - \\ 
   Pred motion  & 37.9 & 52.1 & 0.31 \\ \hline
    \end{tabular}}
    \label{tab:motion}
    \vspace{-5mm}
\end{table}

\noindent{\textbf{Experiments with Different Iterations}}\quad
In Figure~\ref{fig:iteration}, we provide the comparison of modeling object motion and assumed static with different recurrent iterations. Benefiting from the BEV features modeling, the two configurations display almost 2\% mAP improvements in the first iterations. In the later iterations, the improvement in assumed static stops, mainly due to the lack of accurate temporal features. While with more and more accurate temporal features, the model with modeling object motion can progressively improve the detection and motion estimation results. 



\section{Conclusion}
This work proposes a novel framework to better leverage temporal information for camera-only 3D detection by modeling dynamic objects.
We first designed an object-wise local volume to save the computation time and maintain a object-wise representation for motion and detection modeling.
Then we propose a recurrent module to tackle the challenging motion and location modeling problem.
Specifically, we progressively update the motion and location results from the concurrently updated corresponding 3D volume features thereon.
As the object motion and location results are tightly coupled in the recurrent stage, we also demonstrate the framework can naturally achieve 3D object tracking. 
Based on dynamic objects motion, our method achieves state-of-the-art performance on both the nuScenes detection and tracking benchmarks.

{\small
\bibliographystyle{ieee_fullname}
\bibliography{egbib}

\begin{thebibliography}{10}\itemsep=-1pt

\bibitem{Bewley2016_sort}
Alex Bewley, Zongyuan Ge, Lionel Ott, Fabio Ramos, and Ben Upcroft.
\newblock Simple online and realtime tracking.
\newblock In {\em ICIP}, 2016.

\bibitem{brazil2019m3d}
Garrick Brazil and Xiaoming Liu.
\newblock M3d-rpn: Monocular 3d region proposal network for object detection.
\newblock In {\em ICCV}, 2019.

\bibitem{Brazil2020kinematic}
Garrick Brazil, Gerard Pons-Moll, Xiaoming Liu, and Bernt Schiele.
\newblock Kinematic 3d object detection in monocular video.
\newblock In {\em ECCV}, 2020.

\bibitem{nuscene}
Holger Caesar, Varun Bankiti, Alex~H. Lang, Sourabh Vora, Venice~Erin Liong,
  Qiang Xu, Anush Krishnan, Yu Pan, Giancarlo Baldan, and Oscar Beijbom.
\newblock nuscenes: A multimodal dataset for autonomous driving.
\newblock In {\em CVPR}, 2020.

\bibitem{Chen2022PolarDETR}
Shaoyu Chen, , Xinggang Wang, Tianheng Cheng, Qian Zhang, Chang Huang, and
  Wenyu Liu.
\newblock Polar parametrization for vision-based surround-view 3d detection.
\newblock {\em arXiv:2206.10965}, 2022.

\bibitem{fischer2022ccdt}
Tobias Fischer, Yung-Hsu Yang, Suryansh Kumar, Min Sun, and Fisher Yu.
\newblock {CC}-3{DT}: Panoramic 3d object tracking via cross-camera fusion.
\newblock In {\em CORL}, 2022.

\bibitem{Hartley_mvg}
Richard Hartley and Andrew Zisserman.
\newblock {\em Multiple View Geometry in Computer Vision}.
\newblock Cambridge University Press, New York, NY, USA, 2 edition, 2003.

\bibitem{he2017mask}
Kaiming He, Georgia Gkioxari, Piotr Dollár, and Ross Girshick.
\newblock Mask r-cnn.
\newblock In {\em ICCV}, 2017.

\bibitem{he2016deep}
Kaiming He, Xiangyu Zhang, Shaoqing Ren, and Jian Sun.
\newblock Deep residual learning for image recognition.
\newblock In {\em CVPR}, 2016.

\bibitem{Hu2021QD3DT}
Hou-Ning Hu, Yung-Hsu Yang, Tobias Fischer, Trevor Darrell, Fisher Yu, and Min
  Sun.
\newblock Monocular quasi-dense 3d object tracking.
\newblock {\em IEEE Transactions on Pattern Analysis and Machine Intelligence},
  2022.

\bibitem{huang2021bevdet4d}
Junjie Huang and Guan Huang.
\newblock Bevdet4d: Exploit temporal cues in multi-camera 3d object detection.
\newblock {\em arXiv preprint arXiv:2203.17054}, 2022.

\bibitem{huang2021bevdet}
Junjie Huang, Guan Huang, Zheng Zhu, Yun Ye, and Dalong Du.
\newblock Bevdet: High-performance multi-camera 3d object detection in
  bird-eye-view.
\newblock {\em arXiv preprint arXiv:2112.11790}, 2021.

\bibitem{Li_2022_time3d}
Peixuan Li and Jieyu Jin.
\newblock Time3d: End-to-end joint monocular 3d object detection and tracking
  for autonomous driving.
\newblock In {\em CVPR}, June 2022.

\bibitem{li2019multisensor}
Peiliang Li, Siqi Liu, and Shaojie Shen.
\newblock Multi-sensor 3d object box refinement for autonomous driving.
\newblock {\em arXiv preprint arXiv:1909.04942}, 2019.

\bibitem{li2020jst}
Peiliang Li, Jieqi Shi, and Shaojie Shen.
\newblock Joint spatial-temporal optimization for stereo 3d object tracking.
\newblock In {\em CVPR}, 2020.

\bibitem{RTM3D}
Peixuan Li, Huaici Zhao, Pengfei Liu, and Feidao Cao.
\newblock Rtm3d: Real-time monocular 3d detection from object keypoints for
  autonomous driving.
\newblock In {\em ECCV}, 2020.

\bibitem{li2022bevstereo}
Yinhao Li, Han Bao, Zheng Ge, Jinrong Yang, Jianjian Sun, and Zeming Li.
\newblock Bevstereo: Enhancing depth estimation in multi-view 3d object
  detection with dynamic temporal stereo.
\newblock {\em arXiv preprint arXiv:2209.10248}, 2022.

\bibitem{li2022uvtr}
Yanwei Li, Yilun Chen, Xiaojuan Qi, Zeming Li, Jian Sun, and Jiaya Jia.
\newblock Unifying voxel-based representation with transformer for 3d object
  detection.
\newblock In {\em NeurIPS}, 2022.

\bibitem{li2022bevdepth}
Yinhao Li, Zheng Ge, Guanyi Yu, Jinrong Yang, Zengran Wang, Yukang Shi,
  Jianjian Sun, and Zeming Li.
\newblock Bevdepth: Acquisition of reliable depth for multi-view 3d object
  detection.
\newblock {\em arXiv preprint arXiv:2206.10092}, 2022.

\bibitem{li2022bevformer}
Zhiqi Li, Wenhai Wang, Hongyang Li, Enze Xie, Chonghao Sima, Tong Lu, Yu Qiao,
  and Jifeng Dai.
\newblock Bevformer: Learning bird’s-eye-view representation from
  multi-camera images via spatiotemporal transformers.
\newblock In {\em ECCV}, 2022.

\bibitem{lian2022monojsg}
Qing Lian, Peiliang Li, and Xiaozhi Chen.
\newblock Monojsg: Joint semantic and geometric cost volume for monocular 3d
  object detection.
\newblock In {\em CVPR}, 2022.

\bibitem{liu2022petr}
Yingfei Liu, Tiancai Wang, Xiangyu Zhang, and Jian Sun.
\newblock Petr: Position embedding transformation for multi-view 3d object
  detection.
\newblock In {\em ECCV}, 2022.

\bibitem{liu2022petrv2}
Yingfei Liu, Junjie Yan, Fan Jia, Shuailin Li, Aqi Gao, Tiancai Wang, Xiangyu
  Zhang, and Jian Sun.
\newblock Petrv2: A unified framework for 3d perception from multi-camera
  images.
\newblock {\em arXiv:2206.-1256}, 2022.

\bibitem{liu2022convnet}
Zhuang Liu, Hanzi Mao, Chao-Yuan Wu, Christoph Feichtenhofer, Trevor Darrell,
  and Saining Xie.
\newblock A convnet for the 2020s.
\newblock {\em CVPR}, 2022.

\bibitem{autoshape_liu}
Zongdai Liu, Dingfu Zhou, Feixiang Lu, Jin Fang, and Liangjun Zhang.
\newblock Autoshape: Real-time shape-aware monocular 3d object detection.
\newblock In {\em ICCV}, 2021.

\bibitem{lu2022ego3rt}
Jiachen Lu, Zheyuan Zhou, Xiatian Zhu, Hang Xu, and Li Zhang.
\newblock Learning ego 3d representation as ray tracing.
\newblock In {\em ECCV}, 2022.

\bibitem{MonoDLE}
Xinzhu Ma, Yinmin Zhang, Dan Xu, Dongzhan Zhou, Shuai Yi, Haojie Li, and Wanli
  Ouyang.
\newblock Delving into localization errors for monocular 3d object detection.
\newblock In {\em CVPR}, 2021.

\bibitem{Marinello2022CVPR}
Nicola Marinello, Marc Proesmans, and Luc Van~Gool.
\newblock Triplettrack: 3d object tracking using triplet embeddings and lstm.
\newblock In {\em CVPRW}, 2022.

\bibitem{qdtrack_conf}
Jiangmiao Pang, Linlu Qiu, Xia Li, Haofeng Chen, Qi Li, Trevor Darrell, and
  Fisher Yu.
\newblock Quasi-dense similarity learning for multiple object tracking.
\newblock In {\em CVPR}, 2021.

\bibitem{pang2021simpletrack}
Ziqi Pang, Zhichao Li, and Naiyan Wang.
\newblock Simpletrack: Understanding and rethinking 3d multi-object tracking.
\newblock In {\em CVPR}, 2021.

\bibitem{dd3d}
Dennis Park, Rares Ambrus, Vitor Guizilini, Jie Li, and Adrien Gaidon.
\newblock Is pseudo-lidar needed for monocular 3d object detection?
\newblock In {\em ICCV}, 2021.

\bibitem{park2023time}
Jinhyung Park, Chenfeng Xu, Shijia Yang, Kurt Keutzer, Kris~M. Kitani,
  Masayoshi Tomizuka, and Wei Zhan.
\newblock Time will tell: New outlooks and a baseline for temporal multi-view
  3d object detection.
\newblock In {\em The Eleventh International Conference on Learning
  Representations}, 2023.

\bibitem{philion2020lift}
Jonah Philion and Sanja Fidler.
\newblock Lift, splat, shoot: Encoding images from arbitrary camera rigs by
  implicitly unprojecting to 3d.
\newblock In {\em ECCV}, 2020.

\bibitem{roddick2018orthographic}
Thomas Roddick, Alex Kendall, and Roberto Cipolla.
\newblock Orthographic feature transform for monocular 3d object detection.
\newblock {\em arXiv preprint arXiv:1811.08188}, 2018.

\bibitem{Shi2022SRCN3D}
Yining Shi, Jingyan Shen, Yifan Sun, Yunlong Wang, Jiaxin Li, Shiqi Sun, Kun
  Jiang, and Diange Yang.
\newblock Srcn3d: Sparse r-cnn 3d surround-view camera object detection and
  tracking for autonomous driving.
\newblock {\em arXiv:2206.14451}, 2022.

\bibitem{sun2017pwc}
Deqing Sun, Xiaodong Yang, Ming-Yu Liu, and Jan Kautz.
\newblock {PWC-Net}: {CNNs} for optical flow using pyramid, warping, and cost
  volume.
\newblock In {\em CVPR}, 2018.

\bibitem{sun2020disprcnn}
Jiaming Sun, Linghao Chen, Yiming Xie, Siyu Zhang, Qinhong Jiang, Xiaowei Zhou,
  and Hujun Bao.
\newblock Disp r-cnn: Stereo 3d object detection via shape prior guided
  instance disparity estimation.
\newblock In {\em CVPR}, 2020.

\bibitem{slam_review}
Takafumi Taketomi, Hideaki Uchiyama, and Sei Ikeda.
\newblock Visual slam algorithms: a survey from 2010 to 2016.
\newblock {\em IPSJ Transactions on Computer Vision and Applications}, 9, 12
  2017.

\bibitem{tang2018banet}
Chengzhou Tang and Ping Tan.
\newblock {BA}-net: Dense bundle adjustment networks.
\newblock In {\em ICLR}, 2019.

\bibitem{zachary2020raft}
Zachary Teed and Jia Deng.
\newblock Raft: Recurrent all-pairs field transforms for optical flow.
\newblock In {\em ECCV}, 2020.

\bibitem{zachary2021raft3d}
Zachary Teed and Jia Deng.
\newblock Raft-3d: Scene flow using rigid-motion embeddings.
\newblock In {\em CVPR}, 2021.

\bibitem{Wang2019NOCS}
He Wang, Srinath Sridhar, Jingwei Huang, Julien Valentin, Shuran Song, and
  Leonidas~J. Guibas.
\newblock Normalized object coordinate space for category-level 6d object pose
  and size estimation.
\newblock In {\em CVPR}, 2019.

\bibitem{mvdepthnet}
Kaixuan Wang and Shaojie Shen.
\newblock {MVDepthNet}: real-time multiview depth estimation neural network.
\newblock In {\em 3DV}, 2018.

\bibitem{wang2022dfm}
Tai Wang, Jiangmiao Pang, and Dahua Lin.
\newblock Monocular 3d object detection with depth from motion.
\newblock In {\em ECCV}, 2022.

\bibitem{wang2021fcos3d}
Tai Wang, Xinge Zhu, Jiangmiao Pang, and Dahua Lin.
\newblock {FCOS3D}: Fully convolutional one-stage monocular 3d object
  detection.
\newblock In {\em ICCVW}, 2021.

\bibitem{wang2021PGD}
Tai Wang, Xinge Zhu, Jiangmiao Pang, and Dahua Lin.
\newblock Probabilistic and geometric depth: Detecting objects in perspective.
\newblock In {\em CoRL}, 2021.

\bibitem{pseudo_lidar}
Yan Wang, Wei-Lun Chao, Divyansh Garg, Bharath Hariharan, Mark Campbell, and
  Kilian~Q. Weinberger.
\newblock Pseudo-lidar from visual depth estimation: Bridging the gap in 3d
  object detection for autonomous driving.
\newblock In {\em CVPR}, 2019.

\bibitem{wang2019pseudo}
Yan Wang, Wei-Lun Chao, Divyansh Garg, Bharath Hariharan, Mark Campbell, and
  Kilian~Q Weinberger.
\newblock Pseudo-lidar from visual depth estimation: Bridging the gap in 3d
  object detection for autonomous driving.
\newblock In {\em CVPR}, 2019.

\bibitem{detr3d}
Yue Wang, Vitor Guizilini, Tianyuan Zhang, Yilun Wang, Hang Zhao, , and
  Justin~M. Solomon.
\newblock Detr3d: 3d object detection from multi-view images via 3d-to-2d
  queries.
\newblock In {\em CoRL}, 2021.

\bibitem{yan2018second}
Yan Yan, Yuxing Mao, and Bo Li.
\newblock Second: Sparsely embedded convolutional detection.
\newblock {\em Sensors}, 18(10):3337, 2018.

\bibitem{Yang2022QualityTrack}
Jinrong Yang, En Yu, Zeming Li, Xiaoping Li, and Wenbing Tao.
\newblock Quality matters: Embracing quality clues for robust 3d multi-object
  tracking.
\newblock {\em arXiv:2208.10976}, 2022.

\bibitem{yao2018mvsnet}
Yao Yao, Zixin Luo, Shiwei Li, Tian Fang, and Long Quan.
\newblock Mvsnet: Depth inference for unstructured multi-view stereo.
\newblock In {\em ECCV}, 2018.

\bibitem{yao2019recurrent}
Yao Yao, Zixin Luo, Shiwei Li, Tianwei Shen, Tian Fang, and Long Quan.
\newblock Recurrent mvsnet for high-resolution multi-view stereo depth
  inference.
\newblock In {\em CVPR}, 2019.

\bibitem{yin2021center}
Tianwei Yin, Xingyi Zhou, and Philipp Kr{\"a}henb{\"u}hl.
\newblock Center-based 3d object detection and tracking.
\newblock {\em CVPR}, 2021.

\bibitem{you2019pseudo}
Yurong You, Yan Wang, Wei-Lun Chao, Divyansh Garg, Geoff Pleiss, Bharath
  Hariharan, Mark Campbell, and Kilian~Q Weinberger.
\newblock Pseudo-lidar++: Accurate depth for 3d object detection in autonomous
  driving.
\newblock {\em arXiv preprint arXiv:1906.06310}, 2019.

\bibitem{zhang2022mutr3d}
Tianyuan Zhang, Xuanyao Chen, Yue Wang, Yilun Wang, and Hang Zhao.
\newblock Mutr3d: A multi-camera tracking framework via 3d-to-2d queries.
\newblock {\em arXiv preprint arXiv:2205.00613}, 2022.

\bibitem{MonoFlex}
Yunpeng Zhang, Jiwen Lu, and Jie Zhou.
\newblock Objects are different: Flexible monocular 3d object detection.
\newblock In {\em CVPR}, 2021.

\bibitem{zhou2020tracking}
Xingyi Zhou, Vladlen Koltun, and Philipp Kr{\"a}henb{\"u}hl.
\newblock Tracking objects as points.
\newblock In {\em ECCV}, 2020.

\bibitem{zhou2019objects}
Xingyi Zhou, Dequan Wang, and Philipp Kr{\"a}henb{\"u}hl.
\newblock Objects as points.
\newblock In {\em arXiv preprint arXiv:1904.07850}, 2019.

\bibitem{MonoEF}
Yunsong Zhou, Yuan He, Hongzi Zhu, Cheng Wang, Hongyang Li, and Qinhong Jiang.
\newblock Monocular 3d object detection: An extrinsic parameter free approach.
\newblock In {\em CVPR}, 2021.

\end{thebibliography}
}

\clearpage

\twocolumn[{%
\renewcommand\twocolumn[1][]{#1}%
\begin{center}
    \Large
    \textbf{Appendix}
\end{center}}]
\renewcommand\thesection{\Alph{section}}
\setcounter{section}{0}

\section{Implementation Details}
In the main paper, we have introduced our overall multi-camera 3D object detection and tracking framework and the details of the proposed components. In this supplemental section, we present the details of the other basic modules.
\subsection{Network Architecture}
Our framework is built based on BEVDet and BEVDepth, and we follow them to design the basic modules.

\noindent{\textbf{2D Feature Extraction.}}
Given $N$ multi-view images $I \in \mathcal{R}^{N\times W \times H\times 3}$in each frame, we use a shared 2D backbone to extract the corresponding features. We adopt the standard ResNet-50~\cite{he2016deep} as the backbone and initialize it with ImageNet pre-trained weights. Then we adopt a modified Feature Pyramid Network (FPN)~\cite{yan2018second} to extract the multiple-level features and the output 2D features are downsampled with the ratio of $\frac{1}{16}$ with channel size $256$: $F_{pv} \in \mathcal{R}^{\frac{W}{16}\times \frac{H}{16}\times 256}$.

\noindent{\textbf{View Transformation.}}
Our work is the same as BEVDet and BEVDepth that contains a 2D to 3D view transformation module. Specifically, we first leverage a depth prediction head to predict the depth probability for each pixel. Then we lift the 2D features to a 2.5D frustum space via out-product it with the depth probability. The depth probability range is set as $[0m, 60m]$ with grid size $0.5m$. With the 2.5D frustum features, the 3D features for each local volume are obtained via utilizing the camera intrinsic to project the 3D grid back to the frustum and bi-linear sample the corresponding features. As mentioned in the main paper, we aggregate the 3D volume features along the height dimension and obtain the corresponding object-wise BEV features $F^{obj}_{bev} \in \mathcal{R}^{N\times W^{obj} \times H^{obj} \times 256}$, where $W^{obj}$ and $H^{obj}$ are the object features dimension and set as 28 in the main setting.

\noindent{\textbf{RefineNet.}}
Given the object-wise features extracted based on the proposal 3D box and motion, RefineNet takes several convolutional neural networks to extract the object-wise features and estimate the bounding box and motion residual. Specifically, we first adopt an average pooling layer to aggregate the 3D features along the height dimension and obtain the BEV features. Then we filter each object-wise BEV features with 6 basic 2D residual blocks, where each residual block consists of two 2D convolution layers and a skip connection module as in ResNet. The channel size of the residual blocks in the first three layers is 256 and decreases to 64 in the last three layers. Then we aggregate the features along the spatial dimension via average pooling and take 4 layers MLP network to estimate the bounding box and motion residuals.

\subsection{The Tracking Module}
\label{sec:track_supp}
In this section, we provide the details of the tracking module that omit in the paper.
Since DORT can estimate tightly coupled object location and motion, object tracking can be easily achieved via nearest center distances association~\cite{Bewley2016_sort, yin2021center, pang2021simpletrack}. 
Hence, our tracking module is mainly adapted from the previous distance-based object tracker~\cite{Bewley2016_sort, yin2021center, pang2021simpletrack}.  Specifically, the tracking module contains four parts: Pre-processing, Association, Status Update and Life-cycle Management.

\noindent{\textbf{Pre-processing.}}
Given the detection results, the pre-processing stages mainly focus on filtering false negative objects. In our work, we first adopt Non-maximum Suppression to remove the duplicated bounding boxes with the threshold of 0.1 in terms of 3D IoU. Then we filter out the bounding boxes that the confidence threshold is lower than 0.25.

\noindent{\textbf{Association.}}
This stage associates the detection results in the current frame with tracklets in the past frame. Specifically, we first utilize the estimated object motion (velocity) to warp the detection results back to the past frame and then utilize the L2 distances of object centers to compute the similarity between the detected objects and the tracklets. Then we utilize the linear greedy matching strategy to achieve multi-object matching. 

\noindent{\textbf{Status Update.}}
This stage updates the status of the tracklets. For the tracklets that do not match with any bounding boxes, we replace it object center location with the corresponding detection results. For the unmatched objects, we utilize the estimated object velocity to update its object center location. 

\noindent{\textbf{Life-cycle Management.}}
The life-cycle management module controls the ``birth'' and ``depth'' of the tracklets (\textit{i.e.} birth, depth). Specifically, for the unmatched bounding boxes, they will be initialized as new tracklets. For the unmatched tracklets, we remove them when they are consecutive unmatched more than 2 times.

\section{Ablation Studies}

\label{sec:track_exp_supp}
In this section, we provide the additional ablation studies that omit in the main paper.
We will release the code afterward for providing the details of the methods and reproducing the experimental results.
\noindent\textbf{DORT with Different Proposal Detector.}
We first show that DORT is agnostic with different proposal detectors (\textit{e.g.} PGD~\cite{wang2021PGD}, BEVDepth~\cite{li2022bevdepth}). In Table~\ref{tab:dort_proposal_det}, we display the experimental results of DORT with using PGD and BEVDepth as the proposal detectors. 
We can observe that the DORT is insensitive to the proposal detector and can consistently improve BEVDepth. We
Benefiting from the low computation overhead of BEVDepth in the perspective part and the designed local volume, DORT also can achieve a more lightweight pipeline for dynamic object modeling.

\begin{table}[htb]
    \centering
    \caption{Experimental results on the nuScenes validation set. 1 past frame is adopted in the temporal modeling. $^*$ denotes the BEV FLOPS from the proposal detector.}
    \vspace{2mm}
    \tabcolsep2pt
    \resizebox{0.95\linewidth}{!}{\begin{tabular}{c|cc|cc} \hline
\multirow{2}{*}{Method} & \multirow{2}{*}{mAP} & \multirow{2}{*}{NDS} & \multicolumn{2}{c}{Flops} \\
                        &                      &                      & PV          & BEV         \\ \hline
        BEVDepth &35.1 & 47.5 & 120.4  & 94.5 \\
        DORT with PGD & 37.9 & 52.1 &  238.2 & 40.2\\
        DORT with BEVDepth & \textbf{38.1} & \textbf{52.1} & 120.4 & 74.4$^*$+40.2\\ \hline
    \end{tabular}}
    \label{tab:dort_proposal_det}
\end{table}

\noindent\textbf{Tracking with Semantic Embedding or Geometry Distance.}
In this work, DORT achieves 3D object tracking via the nearest centerness association. To have a more comprehensive comparison of the tracking pipeline designed, we further provide the comparison of DORT with using semantic embedding to associate objects. Specifically, we follow previous methods~\cite{Hu2021QD3DT} and adopt the widely-used quasi-dense similarity learning~\cite{qdtrack_conf} to learn the tracking embedding. We extract two kinds of embedding features, one is from the perspective-view (PV) and another is from the bird-eye-view (BEV).  In Table~\ref{tab:track_abla}, we display the tracking results on the nuScenes tracking set. We can observe that DORT with geometry distance association can outperform the embedding-based methods by a large margin. Furthermore, it is also much simpler and more efficient that does not need to maintain an extra object embedding. Besides, the PV embedding is worse than the BEV-based embedding, which may be due to the view change in different cameras. 
\label{sec:track_exp}
\begin{table}[htb]
    \centering
    \caption{Experimental results on the nuScenes validation set. 1 past frame is adopted in the temporal modeling.}
    \vspace{2mm}
    \tabcolsep1pt
   \resizebox{0.95\linewidth}{!}{
    \begin{tabular}{c|ccc} \hline
        Method & AMOTA$\uparrow$ & AMOTP$\downarrow$ & MOTAR$\uparrow$ \\ \hline
        PV-Embedding & 36.8 & 1.412& 44.2 \\
        BEV-Embedding &40.1 &1.356 &46.7 \\
        DORT (Geometry Distance) &\textbf{42.4} &\textbf{1.264} &\textbf{49.2} \\ \hline
    \end{tabular}}
    \label{tab:track_abla}
\end{table}

\section{Theoretical Analysis of Ignoring Object Motion}
\label{sec:analysis}
In the main paper, we have shown that when ignoring object motion, the temporal correspondence would derive a biased depth. In this supplementary, we provide the full details of how ignoring object motion introduces a biased depth.
We denote the camera intrinsic as $K$ and the ego-motion from frame $t_0$ to frame $t_1$ as $T_{t_0\rightarrow t_1}^{ego}$:
\begin{align}
    K = \left[\begin{array}{cccc}
        f & 0 & c_u  \\
        0 & f & c_v \\
        0 & 0 & 1  \\
    \end{array}\right], 
    T_{t_0\rightarrow t_1}^{ego} = 
    \left[
    \begin{array}{cccc}
    1 & 0 & 0 & x^{ego} \\
    0 & 1 & 0 & 0 \\
    0 & 0 & 1 & z^{ego} \\
    \end{array}
    \right].
\end{align}
Here, $f$ is the camera's focal length, and $(c_u, c_v)$ is the camera center coordinates in the image. For simplicity, we assume the ego-motion only contains the translation $(x^{ego}, 0, z^{ego})$ on the horizontal plane. The analysis also can be easily extended to a more complicated case that the motion contains rotation.
Given the multiple-view images, temporal-based methods can utilize photometric or featuremetric similarity to find the correspondence of pixel $p_{t_0} = (u_{t_0}, v_{t_0})$ in the past frame $t_0$ and the pixel $p_{t_1} = (u_{t_1}, v_{t_1})$ in the current frame $t_1$. 

When we ignore the object motion, the depth $z_{t_1}$ of pixel $p_{t_1}$ can be recovered as:
\begin{align}
    \label{eq:nomotion_supp}
    & T_{t_0\rightarrow t_1}^{ego}\cdot \pi(p_{t_0}, K) = \pi(p_{t_1}, K), \nonumber \\
    & z_{t_1}\frac{u_{t_1} + c_u}{f} - x^{ego} = \frac{u_{t_0} + c_u}{f}(z_{t_1} - z^{ego}), \nonumber \\
    & z_{t_1} = \frac{z^{ego}(u_{t_0} - c_u) - fx^{ego}}{u_{t_0} - u_{t_1}}, 
\end{align}
where $\pi$ denotes the projection from 2D image coordinate to 3D camera coordinate.

But as we showed in the main paper, the moving objects occupy large ratios in the driving scenarios. For example, when the object contains the translation $(x^{obj}, 0, z^{obj})$ in the horizontal plane, the object's motion can be represented as 
\begin{align}
    T^{obj}_{i \rightarrow j} = \left[
        \begin{array}{cccc}
            1 & 0 & 0 & x^{obj} \\
            0 & 1 & 0 & 0 \\
            0 & 0 & 1 & z^{obj} \\
        \end{array}
    \right].
\end{align}
With the object motion, the depth $z_{t_1}$ of pixel $p_{t_1}$ is recovered as:
\begin{align}
    \label{eq:motion_supp}
    & T^{obj}_{t_0 \rightarrow t_1}T_{t_0\rightarrow t_1}^{ego} \cdot \pi(p_{t_0}, K) = \pi(p_{t_1}, K), \nonumber \\
    &  z_{t_1}\frac{u_{t_1} + c_u}{f} - x^{ego} - x^{obj} = \frac{u_{t_0} + c_u}{f}(z_{t_1} - z^{ego} - z^{obj}) \nonumber \\
    & \hat{z}_{t_1} = \frac{(z^{ego} + z^{obj})(u_{t_0} - c_u) - f(x^{ego} + x^{obj})}{u_{t_0} - u_{t_1}}.
\end{align} 

From Eq~\eqref{eq:nomotion_supp} and Eq~\eqref{eq:motion_supp}, we can obtain the depth gap for the temporal correspondence with and without considering object motion:
\begin{align}
    \Delta z = \frac{z^{obj} (u_{t_0} - c_u) - fx^{ego}}{u_{t_0} - u_{t_1}}.
    \label{eq:bias_depth_supp}
\end{align}

\begin{figure}
    \centering
    \includegraphics[width=0.4\textwidth]{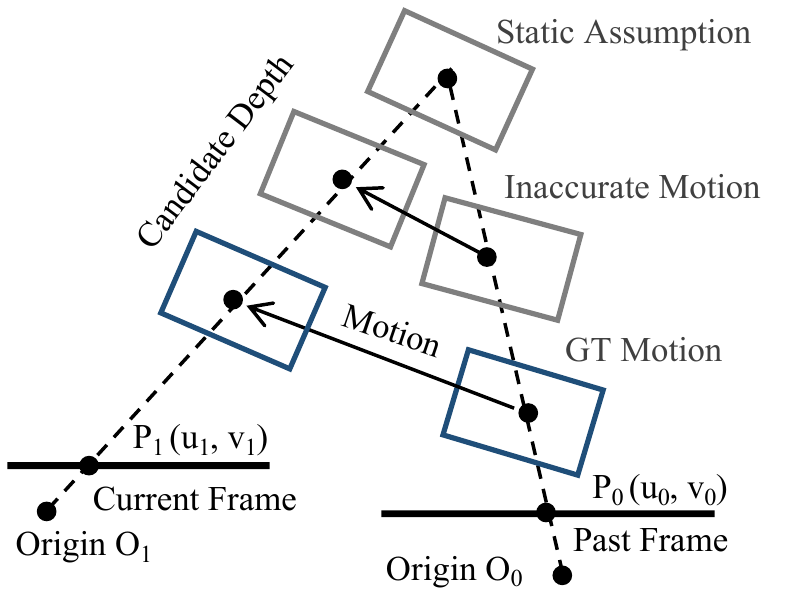}
    \caption{Different object motion can make the same temporal correspondence derive different depth.}
    \label{fig:supp_motion}
\end{figure}

In Figure~\ref{fig:supp_motion}, we also provide a toy example to illustrate that
one temporal correspondence can come from multiple combinations of object depth and motion (\textit{i.e.} inaccurate depth with zero motion and accurate depth and GT motion).
This means that if we inaccurately assume that objects are static across frames, the temporal correspondence would derive a misleading depth.
\subsection{Ill-posed Problem of Simultaneously Estimating 3D Location and Motion}

Although object motion plays a critical role in temporal correspondence, however, it is non-trivial to estimate it from the monocular images.
As shown in Figure~\ref{fig:supp_motion}, the one correspondence can come from infinite combinations of location and motion (the location can be the point in the ray $
 \overrightarrow{O_{t_0}P_{t_0}}$ and $\overrightarrow{O_{t_1}P_{t_1}}$, and the motion can be the line that connects the points.)
Hence, it is an ill-posed problem that simultaneously estimates the 3D location and motion from the monocular images. To alleviate this issue, we leverage the rigid-body assumption for the objects in the driving scenarios and elaborate more temporal frames with constant velocity regularization to further constrain the motion.


\end{document}